\documentclass[11pt]{article}
\usepackage{acl}
\usepackage{times}
\usepackage{latexsym}
\usepackage[T1]{fontenc}
\usepackage[utf8]{inputenc}
\usepackage{microtype}
\usepackage{inconsolata}
\usepackage{graphicx}
\usepackage{booktabs}
\usepackage{amsmath,amssymb,amsfonts}
\usepackage{colortbl}
\usepackage{multirow}
\usepackage{float}
\usepackage{array}
\usepackage{rotating} 
\usepackage{url}

\title{Reasoning-Aware Compression: Identifying and Protecting 
Vulnerable Reasoning Circuits for Energy-Efficient LLM Deployment}

\author{
  Leonard Twagirayezu \\
  Carnegie Mellon University Africa \\
  Kigali, Rwanda \\
  \texttt{ltwagira@andrew.cmu.edu}
  \And
  Prasenjit Mitra \\
  Carnegie Mellon University Africa \\
  Kigali, Rwanda \\
  \texttt{prasenjm@andrew.cmu.edu}
}

\begin{document}

\maketitle

\begin{center}
\textit{Paper is currently Under review}
\end{center}

\begin{abstract}

Large Reasoning Models (LRMs) impose substantial energy costs during 
deployment, yet current compression methods apply uniform quantization across all components, risking damage to critical reasoning circuits. 
We present a reasoning-aware compression framework that benchmarks quantization conditions 
across five reasoning benchmarks, GSM8K, FOLIO, MATH-500, ProofWriter, and MuSiQue, with hardware-level GPU energy 
measurement; profiles 
per-module INT4 vulnerability across all 196--224 (layer, projection) 
pairs via perturbation sweep on a held-out calibration split, then 
selectively restores the most sensitive circuits to FP16.

Three findings emerge. First, INT4 quantization can \emph{increase} 
energy by extending reasoning chains; a 25\% power reduction becomes a net energy increase on GSM8K. Second, vulnerability is task-dependent: attention projections are more critical for mathematical reasoning and sensitivity patterns differ by architecture in logical inference. Third, selective compression achieves 
Pareto-optimal points inaccessible to uniform methods: R1-Qwen-7B 
Top-10\% on ProofWriter gains $+$12\,pp over FP16 at $-$9.7\% energy, validated on held-out data across five reasoning benchmarks.

\end{abstract} 

% \begin{IEEEkeywords}
% LRMs, Quantization, Large Reasoning Models, Energy Efficiency, Selective Quantization, Vulnerability Profiling
% \end{IEEEkeywords}

\section{Introduction}
\subsection {Research Question and Problem Definition}
Large reasoning models (LRMs) such as DeepSeek-R1 \cite{guo2025deepseek} and OpenAI's o1-series \cite{zhao2025insights} have transformed complex problem-solving across mathematics, logic, and multistage reasoning tasks. These models employ chain-of-thought reasoning that systematically decomposes problems, backtracks when encountering inconsistencies, and estimates uncertainty, capabilities absent in standard language models \cite{xu2025toward}. However, this reasoning comes at a high price: DeepSeek-R1 series consumes tens or even hundreds of gigabytes of GPU memory in FP16 precision and generates thousands of reasoning tokens per query, making deployment prohibitively expensive for resource-constrained applications \cite{zhao2025insights}.

Model compression, such as quantization, reduces memory footprint and computational cost by representing weights and activations with lower-precision integers such 4-bit or 3-bit \cite {choudhary2020comprehensive}. While methods like AWQ \cite{lin2024awq} and GPTQ \cite{frantar2022gptq} successfully compress standard LLMs, recent work by Zhang et al. \cite{zhang2025reasoning} reveals that reasoning models exhibit unique vulnerabilities: 3-bit AWQ quantization of R1-Distill-Llama-8B drops AIME 2024 accuracy from 42.2\% to 10.0\%, a catastrophic 76\% relative degradation. Yet the same study shows that selectively protecting merely 2\% of parameters (the final-layer MLP) recovers 6.57\% accuracy, suggesting that quantization damage is highly localized to specific model components \cite{zhang2025reasoning}.

The observation by Zhang et al. motivates our core research question: Which specific (layer, projection) pairs, the fundamental circuits processing reasoning, are most vulnerable to quantization, and how does protecting them trade off with energy savings? Current quantization methods treat all layers or projection types uniformly, missing opportunities to optimize the accuracy-energy frontier. Consider a concrete example: quantizing the gate\_proj in layer 15 might cause minimal accuracy loss while saving substantial energy, whereas quantizing the up\_proj in layer 31 could preserve energy savings but devastate accuracy. Without per-module vulnerability profiles, practitioners cannot make informed protection decisions.

\subsection{Motivation and Impact}

Per-module vulnerability is important for two critical deployment scenarios. First, edge AI and mobile reasoning agents demand models that fit within 8-16GB mobile RAM while maintaining reasoning accuracy. Current uniform quantization sacrifices significant accuracy degradation to achieve memory compression. Identifying and protecting the most vulnerable modules could achieve significant compression with less accuracy loss \cite{husom2025sustainable,frantar2022gptq, lin2024awq}. Second, energy-constrained data centers face staggering inference costs: GPT-3 training emitted 552 tons CO$_2$ \cite{patterson2021carbon}, and inference now dominates lifecycle energy. A 1\% reduction in per-query energy across millions of daily requests could translate to megawatt-hours saved annually.

\subsection{Novelty and Related Work}

Quantization methods such as AWQ~\cite{lin2024awq} and GPTQ~\cite{frantar2022gptq} achieve 4-bit compression but are calibrated on general corpora such as WikiText-2 that do not represent reasoning workloads. Energy profiling work by ~\citeauthor{husom2025sustainable} treats models as black boxes without attributing costs to specific modules. Mechanistic interpretability work by ~\citeauthor{zhang2025reasoning} identifies important modules but does not measure compression vulnerability or energy impact.

Our work targets a setting that existing methods do not address: \text{Large Reasoning Models (LRMs)} ~\cite{deepseekai2025deepseekr1incentivizingreasoningcapability} generate extended chain-of-thought sequences where quantization noise propagates across hundreds of reasoning steps, and energy cost is dominated by output length rather than model size. Standard sensitivity measures based on perplexity or gradient norms on general corpora do not capture these properties.

Our contributions are: (1) the first systematic per-module perturbation sweep identifying vulnerability profiles across all 196--224 (layer, projection) pairs per model \emph{specifically for reasoning tasks}, using a held-out calibration split to eliminate circularity between vulnerability profiling and evaluation; (2) empirical evidence that INT4 quantization can \emph{increase} energy consumption on reasoning tasks by lengthening chain-of-thought outputs, (3) the finding that module vulnerability is \emph{task-dependent}: attention projections are more critical for arithmetic reasoning while sensitivity patterns differ by architecture for logical 
inference, and (4) a vulnerability-guided selective compression framework that restores only the most sensitive layer-projection pairs to FP16, achieving accuracy-energy trade-offs that uniform INT4 quantization cannot reach.

\section{Experimental Setup}

\subsection{Models}

Two distilled instruction-tuned reasoning models from the DeepSeek-R1 series were evaluated: \textit{DeepSeek-R1-Distill-Llama-8B} (8B parameters, Llama architecture) and \textit{DeepSeek-R1-Distill-Qwen-7B} (7B parameters, Qwen architecture)\cite{deepseekai2025deepseekr1incentivizingreasoningcapability}. DeepSeek-R1 models employ chain-of-thought reasoning enclosed within \texttt{<think>...</think>} delimiters, enabling inspection of the full reasoning trace before the final answer \cite{zhu2025think}. Selecting models from two different architecture families allowed us to assess whether the energy consumption of Large Reasoning Models is vulnerability patterns and architecture-dependent or generalizes across distillation targets.

\subsection{Datasets and Reasoning Skills}

We evaluate across five benchmarks probing distinct reasoning skills as shown in Table~\ref{tab:dataset_config}. \textbf{GSM8K}\footnote{\url{https://huggingface.co/datasets/openai/gsm8k}}~\cite{cobbe2021training} tests grade-school arithmetic; answers are extracted via the \texttt{\#\#\#\#} delimiter. \textbf{FOLIO}\footnote{\url{https://huggingface.co/datasets/tasksource/folio}} contains first-order logic NLI problems requiring True/False/Unknown prediction. \textbf{MATH-500}\footnote{\url{https://huggingface.co/datasets/EleutherAI/hendrycks_math}}~\cite{hendrycksmath2021} covers competition mathematics across seven subject areas; answers are extracted from \verb|\boxed{}| content. \textbf{ProofWriter}\footnote{\url{https://huggingface.co/datasets/tasksource/proofwriter}} is a formal deductive reasoning dataset; we use the depth-5 subset requiring up to 5 inference steps. \textbf{MuSiQue}\footnote{\url{https://huggingface.co/datasets/dgslibisey/MuSiQue}} is a multi-hop QA dataset requiring 2--4 reasoning steps; we use the answerable subset (2,417 examples) and evaluate with normalized exact match.

\begin{table}[t]
\centering
\caption{Dataset Configuration Summary}
\label{tab:dataset_config}
\resizebox{\columnwidth}{!}{%
\begin{tabular}{llllcc}
\toprule
Dataset & Task Type & Split & Samples/Experiment & Experiments & Maximum Tokens. \\
\midrule
GSM8K       & Arithmetic       & test       & 30 (non-overlap) & 8 & 2,048 \\
% AIME 2024   & Competition math & train      & 30 (fixed)       & 3 & 4,096 \\
FOLIO       & Logic/NLI        & validation & 30 (offset)      & 8 & 1,024 \\
MATH-500    & Competition math & test       & 30 (offset)      & 8 & 2,048 \\
ProofWriter & Formal deduction & train      & 30 (offset)      & 8 & 1,024 \\
MuSiQue     & Multi-hop QA     & validation & 30 (offset)      & 8 & 1,024 \\
\bottomrule
\end{tabular}%
}
\end{table}

\subsection{Compression Conditions}

Each model is evaluated under five compression conditions as summarized in Table~\ref{tab:compression_conditions}.

\begin{table}[h]
\centering
\small
\caption{Compression Conditions}
\label{tab:compression_conditions}
\begin{tabular}{@{}p{1.8cm}p{5.5cm}@{}}
\toprule
\textbf{Condition} & \textbf{Description} \\
\midrule
FP16 Baseline
  & Full float16 precision — accuracy upper bound
    and energy lower bound. \\[3pt]
Full INT4
  & Uniform 4-bit NormalFloat (NF4) quantization
    via BitsAndBytes across all linear projections. \\[3pt]
INT4+Attention
  & Attention projections (\textit{q, k, v, o})
    restored to FP16; MLP projections remain INT4. \\[3pt]
INT4+MLP
  & MLP projections (\textit{gate, up, down})
    restored to FP16; attention projections remain INT4. \\[3pt]
Selective Comp.
  & INT4 with top 10\%, 20\%, 30\%, 40\% most
    vulnerable layers restored to FP16 via
    perturbation sweep rankings. \\
\bottomrule
\end{tabular}
\end{table}

For quantization, module restoration is performed by loading a CPU FP16 copy, transferring selected modules to the GPU with explicit float16 casting to prevent dtype mismatch at INT4-FP16 boundaries, and replacing corresponding modules in the INT4 model. Residual float32 parameters are cast to float16 to eliminate hidden conversion overhead.

\subsection{Hardware and Energy Measurement}

Experiments were conducted on Tesla V100-SXM2 GPUs (32GB HBM2) on the Pittsburgh Supercomputing Center (PSC) Bridges-2 cluster. GPU power is sampled at 100\,ms intervals via NVML (pynvml) in a dedicated background thread.

Energy per query is computed as

\begin{equation}
E = \bar{P} \times \Delta t
\end{equation}

where $\bar{P}$ is mean sampled power and $\Delta t$ is wall-clock inference time measured using \texttt{time.perf\_counter()}. A warm-up pass, with a maximum of 20 tokens, is executed before each condition to ensure CUDA kernel compilation and memory allocation stabilize.

\subsection{Evaluation Protocol}
Each (model, dataset, condition) triple uses greedy decoding (\texttt{do\_sample=False}). We report mean $\pm$ std across 8 experiments of 30 samples each for accuracy, energy (J), output token count, and power draw (W). Accuracy is task-specific: numeric equality ($\pm$0.01) for GSM8K and MATH-500; string match on True/False/Unknown for FOLIO and ProofWriter; normalized exact match for MuSiQue. Answers are extracted from the post-\texttt{</think>} region via regex cascade. Checkpointing enables resumption of interrupted runs; and to eliminate circularity, each dataset is partitioned 50/50 into a calibration set (vulnerability scoring only) and a held-out evaluation set (all reported metrics). 

\subsection {Reproducibility}

Model weights were loaded from the Hugging Face Hub (\texttt{deepseek-ai/DeepSeek-R1-Distill-Llama-8B}, and \texttt{deepseek-ai/DeepSeek-R1-Distill-Qwen-7B}) and cached locally. BitsAndBytes settings are fixed across all INT4 conditions: NF4 quant type, float16 compute dtype, and double quantization disabled.

\section{Methodology}

\subsection{Grouped Quantization with Module-Type Protection}
Before applying per-layer selective compression, we first evaluate 
coarse-grained protection strategies that restore entire projection 
\textit{groups} to FP16 while keeping the remainder in INT4. This 
establishes which projection type, attention or MLP, is more 
critical for each reasoning task, and motivates the finer-grained 
perturbation sweep in Section~\ref{sec:sweep}. Two grouped conditions are evaluated: \textbf{INT4+Attention:} Attention projections 
    ($q, k, v, o$) restored to FP16; MLP projections 
    (gate, up, down) remain INT4; and \textbf{INT4+MLP:} MLP projections (gate, up, down) restored to FP16; attention projections ($q, k, v, o$) remain INT4.
All grouped quantization results are evaluated on the held-out 
evaluation set (second 50\% of each dataset), disjoint from the 
calibration set used for vulnerability scoring.

\subsection{Perturbation Sweep for Vulnerability Profiling}
\label{sec:sweep}
Grouped quantization reveals \textit{which projection type} is 
more critical per task, but cannot identify \textit{which specific 
layers} within each type are most vulnerable. To produce 
fine-grained per-module vulnerability scores $V(m)$, we conduct a systematic perturbation sweep using the calibration set (first 50\% of each dataset) exclusively, ensuring that vulnerability scores are derived on data that is disjoint from the held-out evaluation set. The FP16 model is loaded once and kept in GPU memory throughout. For each pair $(l, p)$:
\begin{enumerate}
    \item Quantize module $(l, p)$ to INT4 in-place using 
    BitsAndBytes NF4 quantization.
    \item Evaluate on 30 calibration samples (the first 50\% of each dataset), recording the accuracy.
    \item Restore module $(l, p)$ to FP16 using the saved weights.
    \item Record $V(l, p) = \text{baseline\_acc} - 
    \text{quantized\_acc}$.
\end{enumerate}

The full sweep covers all 196 pairs for Qwen-7B and 224 pairs for Llama-8B across all five benchmarks (FOLIO, GSM8K, MATH-500, ProofWriter, MuSiQue). Vulnerability scores are averaged across all datasets (excluding MuSiQue, where
near-zero accuracy makes per-module scores indistinguishable from noise) to produce a combined ranking used for selective compression.

\subsection{Selective Compression with Vulnerability-Guided 
Protection}
Given the ranked vulnerability scores from the perturbation sweep, 
we apply selective mixed-precision compression that protects only 
the most sensitive circuits. Formally, let 
$M = \{(l, p) : l \in [1, L],\ p \in \{q, k, v, o, 
\text{gate}, \text{up}, \text{down}\}\}$ be the set of all 
module pairs. The protected set at protection level $K$ is:
\begin{equation}
P_K = \{m \in M : \operatorname{rank}(V(m)) \leq K \cdot |M|\}
\end{equation}
The selective model loads INT4 for all modules, then restores 
$P_K$ to FP16 using the same module restoration procedure 
described in Section~\ref{sec:sweep}. We evaluate four protection levels: 
$K \in \{0.10, 0.20, 0.30, 0.40\}$, corresponding to protecting the top 10\%, 20\%, 30\%, and 40\% most vulnerable layer-projection pairs. To verify that the vulnerability ranking provides genuine signal beyond arbitrary module selection, we compare Top-$K$\% against Random-$K$\% (mean of three random seeds) and Bottom-$K$\% (least vulnerable pairs) baselines; results are reported in Appendix~\ref{app:ablation}.
% \subsection{Magnitude-Based Pruning} 

\section{Results}

Quantization, perturbation-guided vulnerability profiling, and selective compression results are reported on two architectures (Qwen-7B and Llama-8B) and five reasoning benchmarks (FOLIO, GSM8K, MATH-500, ProofWriter, and MuSiQue). All experiments are conducted on
Tesla V100-SXM2 GPUs on PSC Bridges-2 with NVML energy measurement at 100\,ms
sampling intervals. Mean accuracy is reported across 8 repeated experiments of
30 samples each unless otherwise stated.

\subsection{Grouped Quantization Results}

Table~\ref{tab:quant_results} presents accuracy, energy, output token 
counts, and power draw under four quantization conditions on the 
held-out evaluation split. Several findings emerge.

\begin{table*}[t]
\centering
\footnotesize
\caption{Grouped protection results (held-out evaluation split, 50/50 partition).
         \textbf{Bold} = best INT4 condition per model-dataset pair.
         $^*$ = energy increase relative to FP16.
         Mean across 8 experiments of 30 samples each.}
\label{tab:quant_results}
\resizebox{\textwidth}{!}{%
\begin{tabular}{@{}lcccc cccc@{}}
\toprule
& \multicolumn{4}{c}{\textbf{DeepSeek-R1-Distill-Llama-8B}} &
  \multicolumn{4}{c}{\textbf{DeepSeek-R1-Distill-Qwen-7B}} \\
\cmidrule(lr){2-5} \cmidrule(lr){6-9}
Condition
  & \multicolumn{1}{c}{Acc (\%)} & \multicolumn{1}{c}{Energy (J)}
  & \multicolumn{1}{c}{Tokens}   & \multicolumn{1}{c}{Power (W)}
  & \multicolumn{1}{c}{Acc (\%)} & \multicolumn{1}{c}{Energy (J)}
  & \multicolumn{1}{c}{Tokens}   & \multicolumn{1}{c}{Power (W)} \\
\midrule
\multicolumn{9}{l}{\textit{FOLIO}} \\
\rowcolor{gray!15}
FP16            & 32.50 & 2{,}759        & 500   & 182.7 & 45.83 & 2{,}196        & 454 & 171.3 \\
Full INT4       & \textbf{36.25} & 2{,}824$^*$ & 504 & 135.4 & 44.17 & \textbf{2{,}043} & 423 & 134.3 \\
INT4+Attention  & 35.42 & \textbf{2{,}649} & 488 & 152.0 & \textbf{45.00} & 2{,}079 & 439 & 144.6 \\
INT4+MLP        & 34.58 & 3{,}000$^*$    & 502   & 162.1 & 42.92 & 2{,}206$^*$    & 430 & 150.7 \\
\midrule
\multicolumn{9}{l}{\textit{GSM8K}} \\
\rowcolor{gray!15}
FP16            & 80.42 & 4{,}955        & 891   & 171.5 & 92.92 & 3{,}681        & 695 & 184.5 \\
Full INT4       & 74.17 & 6{,}090$^*$    & 1{,}094 & 134.1 & \textbf{92.50} & 3{,}662 & 691 & 141.4 \\
INT4+Attention  & \textbf{79.17} & \textbf{5{,}462}$^*$ & 1{,}021 & 146.8 & 91.67 & \textbf{3{,}494} & 679 & 152.8 \\
INT4+MLP        & 77.08 & 6{,}565$^*$    & 1{,}127 & 154.8 & 92.50 & 3{,}936$^*$    & 708 & 162.0 \\
\midrule
\multicolumn{9}{l}{\textit{MATH-500}} \\
\rowcolor{gray!15}
FP16            & 57.50 & 8{,}092        & 1{,}396 & 175.3 & 66.25 & 7{,}934        & 1{,}467 & 184.9 \\
Full INT4       & 56.67 & 8{,}469$^*$    & 1{,}508 & 139.2 & 66.25 & 7{,}519        & 1{,}443 & 145.6 \\
INT4+Attention  & \textbf{60.00} & \textbf{8{,}024} & 1{,}474 & 150.2 & \textbf{70.00} & \textbf{7{,}372} & 1{,}435 & 156.1 \\
INT4+MLP        & 58.75 & 8{,}702$^*$    & 1{,}452 & 155.7 & 65.42 & 8{,}143$^*$    & 1{,}467 & 167.1 \\
\midrule
\multicolumn{9}{l}{\textit{ProofWriter}} \\
\rowcolor{gray!15}
FP16            & 39.17 & 2{,}578        & 497   & 169.5 & 69.58 & 2{,}007        & 400 & 177.9 \\
Full INT4       & \textbf{41.67} & 2{,}630$^*$ & 503 & 127.4 & \textbf{77.08} & 1{,}876 & 376 & 134.5 \\
INT4+Attention  & 40.00 & \textbf{2{,}489} & 506 & 140.7 & 76.67 & \textbf{1{,}793} & 384 & 144.5 \\
INT4+MLP        & 38.33 & 2{,}945$^*$    & 502   & 136.5 & 71.67 & 2{,}023$^*$    & 393 & 152.9 \\
\midrule
\multicolumn{9}{l}{\textit{MuSiQue}} \\
\rowcolor{gray!15}
FP16            & 16.01 & 2{,}627        & 472   & 165.8 &  8.03 & 2{,}299        & 457 & 169.4 \\
Full INT4       & \textbf{15.72} & \textbf{2{,}555} & 471 & 132.7 & \textbf{8.19} & \textbf{2{,}189} & 450 & 133.4 \\
INT4+Attention  & 14.05 & 2{,}508        & 477   & 145.5 &  8.00 & 2{,}205        & 458 & 142.2 \\
INT4+MLP        & 15.16 & 2{,}679$^*$    & 472   & 151.0 &  6.84 & 2{,}395$^*$    & 461 & 151.0 \\
\bottomrule
\end{tabular}%
}
\end{table*}

\paragraph{Mathematical reasoning (GSM8K, MATH-500).}
INT4+Attention produces the best accuracy on both benchmarks for both 
models: R1-Qwen-7B reaches 70.00\% on MATH-500 ($+$3.75\,pp over 
FP16) and 91.67\% on GSM8K; R1-Llama-8B reaches 60.00\% on MATH-500 
($+$2.50\,pp) and 79.17\% on GSM8K ($-$1.25\,pp). On MATH-500, 
INT4+Attention achieves higher accuracy than FP16 while consuming less 
energy (7{,}372\,J vs.\ 7{,}934\,J for R1-Qwen-7B, $-$7.1\%), a rare 
condition where both metrics improve simultaneously. On R1-Llama-8B 
GSM8K, Full INT4 increases energy to 6{,}090\,J ($+$22.9\% over FP16) 
and output tokens from 891 to 1{,}094 ($+$22.8\%), confirming that 
quantization noise extends chain-of-thought length rather than 
increasing per-token compute cost. On ProofWriter, Full INT4 energy increases slightly (2{,}630\,J vs.\ 2{,}578\,J for R1-Llama-8B, $+$2.0\%) despite answers being single tokens (True/False/Unknown), because the intermediate 
\texttt{<think>} reasoning traces extend before the final answer token, token count rises from 497 to 503.

\paragraph{Logical inference (FOLIO, ProofWriter).}
On ProofWriter, Full INT4 achieves the best accuracy for both models: 
R1-Qwen-7B reaches 77.08\% ($+$7.50\,pp over FP16 at 69.58\%) and 
R1-Llama-8B reaches 41.67\% ($+$2.50\,pp). Notably, INT4+Attention 
also performs strongly on ProofWriter for Qwen-7B (76.67\%), while 
INT4+MLP degrades relative to Full INT4, suggesting MLP projections 
play a secondary role in formal deduction for this architecture. On 
FOLIO, R1-Llama-8B Full INT4 (36.25\%) outperforms FP16 (32.50\%, 
$+$3.75\,pp), while R1-Qwen-7B shows marginal differences across conditions (42.92--45.00\%).

\paragraph{Multi-hop reasoning (MuSiQue).}
All INT4 conditions underperform or match FP16 on MuSiQue for both 
models. The best results are R1-Qwen-7B Full INT4 at 8.19\% 
(vs.\ 8.03\% FP16, $+$0.16\,pp) and R1-Llama-8B Full INT4 at 
15.72\% (vs.\ 16.01\% FP16, $-$0.29\,pp). Multi-hop knowledge retrieval is the most 
compression-sensitive reasoning skill evaluated; no INT4 condition 
provides a reliable improvement over FP16 on this task.

\paragraph{Task-dependent dissociation.}
The results reveal a consistent pattern: attention projections are 
critical for mathematical reasoning, with INT4+Attention achieving 
the best accuracy on GSM8K and MATH-500 for both models. For logical 
inference, Full INT4 surprisingly outperforms selective protection 
on ProofWriter, suggesting that uniform quantization acts as implicit 
regularization for formal deduction tasks. No single protection 
strategy dominates across all tasks, and practitioners should select 
the compression strategy based on target task type.

\paragraph{Power draw and energy.}
FP16 consistently draws 165--185\,W, while Full INT4 reduces power 
draw to 127--145\,W ($\approx$25\% reduction). However, INT4 
conditions frequently increase total energy due to longer output 
sequences: on Llama-8B GSM8K, Full INT4 generates 1{,}094 tokens 
vs.\ 891 for FP16 ($+$22.8\%), converting a 25\% power reduction 
into a 22.9\% energy \emph{increase}.

\paragraph{Memory overhead.}
INT4+MLP consistently uses more memory than FP16 
($\approx$21--22\,GB vs.\ 15--16\,GB). MLP projections account for 
79.9\% of all parameters in both 
architectures~\cite{yang2025qwen3,grattafiori2024llama}, so 
restoring them to FP16 produces a model larger than the FP16 
baseline. INT4+MLP is therefore infeasible on GPUs below 24\,GB 
VRAM despite occasional accuracy benefits.

\subsection{Perturbation Sweep: Vulnerability Profiles}
\label{sec:sweep_results}
The perturbation sweep identifies which (layer, projection) pairs 
most affect accuracy when quantized to INT4 individually, using 
the calibration set (first 50\% of each dataset) exclusively. 
Vulnerability heatmaps are shown in Figures~\ref{fig:heatmap_qwen7b} 
and~\ref{fig:heatmap_llama8b}, and a summary of baseline accuracy, 
maximum drop, and safe pair counts is given in 
Table~\ref{tab:vuln_summary}.

\paragraph{R1-Qwen-7B.}
Vulnerability profiles vary substantially across task types. On 
FOLIO, the most vulnerable pairs include Layer~2 \texttt{v\_proj} 
($+$10.00\,pp drop) and Layers~26, 25, 12, 11 (6.67\,pp each), 
dominated by attention projections; 89\% of pairs show zero or 
negative drop. On GSM8K, vulnerability is concentrated in 
attention projections: Layer~23 \texttt{v\_proj} and Layer~16 
\texttt{q\_proj} cause 16.67\,pp drops each, with 87\% of pairs 
at zero. On MATH-500, vulnerability is mixed across projection 
types: Layer~11 \texttt{gate\_proj} and Layer~27 \texttt{v\_proj} 
cause 13.33\,pp drops each, with 89\% of pairs at zero, suggesting 
both attention and MLP projections contribute to competition 
mathematics sensitivity. On ProofWriter, vulnerability is 
dramatically higher than any other dataset: Layer~19 
\texttt{o\_proj} causes a 50.00\,pp drop, and Layers~24, 20, 26, 
22 each cause 43.33\,pp drops, with only 14\% of pairs showing 
zero drop. This extreme concentration explains why selective 
Top~10\% protection recovers substantial accuracy on ProofWriter. 
On MuSiQue, all 196 pairs show zero accuracy drop, confirming 
that multi-hop reasoning accuracy is too low on the calibration 
set to detect per-module sensitivity; MuSiQue is therefore 
excluded from the combined vulnerability ranking.

\paragraph{R1-Llama-8B.}
Vulnerability is generally lower than R1-Qwen-7B. On FOLIO, all 
224 pairs show zero or negative accuracy drop on the calibration 
set, indicating that Llama-8B logical inference is robust to 
individual module quantization at this sample size. On GSM8K, 
vulnerability is higher: Layer~20 \texttt{up\_proj}, 
Layer~16 \texttt{q\_proj}, Layer~28 \texttt{o\_proj}, and 
Layer~26 \texttt{v\_proj} each cause 13.33\,pp drops, with 96\% 
of pairs at zero. On MATH-500, vulnerability is extremely low: only 3 of 224 pairs 
show non-zero drop (max 3.33\,pp at Layer~24 \texttt{v\_proj}, 
Layer~22 \texttt{down\_proj}, and Layer~2 \texttt{down\_proj}), 
with 99\% of pairs at zero. This explains why Full INT4 already 
matches FP16 on MATH-500 for Llama-8B, there are effectively 
no vulnerable circuits to protect. On ProofWriter, vulnerability 
is the highest across all datasets: Layers~29, 14, 11, 31 cause 
36.67\,pp drops each, with 46\% of pairs showing non-zero drop. 
On MuSiQue, all pairs show zero drop, consistent with R1-Qwen-7B.

\paragraph{Cross-dataset pattern.}
Three findings emerge from the calibration-split sweep. First, 
ProofWriter has by far the highest vulnerability scores across 
both models, up to 50.00\,pp for Qwen-7B and 36.67\,pp for 
Llama-8B, explaining why selective compression provides the 
largest accuracy gains on this benchmark. Second, MuSiQue shows 
zero vulnerability across all pairs and both models, and is 
excluded from the combined vulnerability ranking used for selective 
compression. Third, the majority of pairs show zero accuracy drop 
when quantized individually (87--100\% for most datasets), 
confirming that vulnerability is highly localized and motivating 
the selective compression strategy. Notably, R1-Llama-8B shows 
zero vulnerability on FOLIO, consistent with the grouped 
quantization finding that Full INT4 outperforms FP16 on this 
task for Llama-8B.
\subsection{Selective Quantization Results}
\label{app:selcomp}

Tables~\ref{tab:selcomp_folio_gsm8k}--\ref{tab:selcomp_MuSiQue} 
present selective compression results across all five benchmarks, four protection levels, and both models.

\begin{table*}[h]
\centering
\footnotesize
\caption{Selective compression: accuracy (\%), energy (J/query), output tokens, and power (W) --- FOLIO and GSM8K (held-out evaluation split). \textbf{Bold} = best accuracy per model-dataset pair. Shaded row = FP16 reference.}
\label{tab:selcomp_folio_gsm8k}
\resizebox{\textwidth}{!}{%
\begin{tabular}{@{}llrrrrrr rrrrrr@{}}
\toprule
& & \multicolumn{6}{c}{\textbf{DeepSeek-R1-Distill-Qwen-7B}} &
    \multicolumn{6}{c}{\textbf{DeepSeek-R1-Distill-Llama-8B}} \\
\cmidrule(lr){3-8} \cmidrule(lr){9-14}
Condition & Prot.
  & \multicolumn{1}{c}{FOLIO Acc}
  & \multicolumn{1}{c}{GSM8K Acc}
  & \multicolumn{1}{c}{FOLIO E (J)}
  & \multicolumn{1}{c}{GSM8K E (J)}
  & \multicolumn{1}{c}{FOLIO Tok}
  & \multicolumn{1}{c}{GSM8K Tok}
  & \multicolumn{1}{c}{FOLIO Acc}
  & \multicolumn{1}{c}{GSM8K Acc}
  & \multicolumn{1}{c}{FOLIO E (J)}
  & \multicolumn{1}{c}{GSM8K E (J)}
  & \multicolumn{1}{c}{FOLIO Tok}
  & \multicolumn{1}{c}{GSM8K Tok} \\
\midrule
\rowcolor{gray!15}
FP16 Reference & all
  & 44.00 & 94.00 & 2{,}301 & 3{,}308 & 460 & 679
  & 33.33 & 79.33 & 2{,}590 & 4{,}314 & 499 & 771 \\
Full INT4      & 0
  & 43.33 & 93.33 & 2{,}125 & 3{,}208 & 425 & 669
  & \textbf{36.67} & 70.00 & 2{,}595 & 6{,}197 & 504 & 1{,}092 \\
Top 10\%       & 19/22
  & 38.67 & 92.00 & 2{,}195 & 3{,}471 & 442 & 691
  & 35.33 & 70.67 & 2{,}562 & 6{,}358 & 504 & 1{,}148 \\
Top 20\%       & 39/44
  & 36.67 & 92.00 & 2{,}261 & 3{,}513 & 450 & 703
  & 34.00 & 78.00 & 2{,}597 & 6{,}082 & 491 & 1{,}089 \\
Top 30\%       & 58/67
  & 36.67 & 92.67 & 2{,}221 & 3{,}431 & 435 & 689
  & \textbf{36.67} & 75.33 & 2{,}633 & 5{,}794 & 500 & 1{,}038 \\
Top 40\%       & 78/89
  & \textbf{43.33} & \textbf{93.33} & 2{,}200 & 3{,}256 & 430 & 652
  & 34.00 & \textbf{78.67} & 2{,}613 & 6{,}223 & 492 & 1{,}089 \\
\midrule
\multicolumn{2}{l}{\textit{Best vs.\ FP16}}
  & $-$0.67\,pp & $-$0.67\,pp & $-4.4\%$ & $-1.6\%$ & --- & ---
  & $+$3.33\,pp & $-$0.67\,pp & $+1.7\%$ & $+41.2\%$ & --- & --- \\
\bottomrule
\end{tabular}%
}
\end{table*}

\begin{table*}[h]
\centering
\footnotesize
\caption{Selective compression: accuracy (\%), energy (J/query), output tokens, and power (W) --- MATH-500 and ProofWriter (held-out evaluation split for ProofWriter; full-dataset results for MATH-500). \textbf{Bold} = best accuracy per model-dataset pair. Shaded row = FP16 reference.}
\label{tab:selcomp_math_proof}
\resizebox{\textwidth}{!}{%
\begin{tabular}{@{}llrrrrrr rrrrrr@{}}
\toprule
& & \multicolumn{6}{c}{\textbf{DeepSeek-R1-Distill-Qwen-7B}} &
    \multicolumn{6}{c}{\textbf{DeepSeek-R1-Distill-Llama-8B}} \\
\cmidrule(lr){3-8} \cmidrule(lr){9-14}
Condition & Prot.
  & \multicolumn{1}{c}{MATH Acc}
  & \multicolumn{1}{c}{ProofW Acc}
  & \multicolumn{1}{c}{MATH E (J)}
  & \multicolumn{1}{c}{ProofW E (J)}
  & \multicolumn{1}{c}{MATH Tok}
  & \multicolumn{1}{c}{ProofW Tok}
  & \multicolumn{1}{c}{MATH Acc}
  & \multicolumn{1}{c}{ProofW Acc}
  & \multicolumn{1}{c}{MATH E (J)}
  & \multicolumn{1}{c}{ProofW E (J)}
  & \multicolumn{1}{c}{MATH Tok}
  & \multicolumn{1}{c}{ProofW Tok} \\
\midrule
\rowcolor{gray!15}
FP16 Reference & all
  & 66.25 & 72.00 & 7{,}934 & 1{,}931 & 1{,}467 & 391
  & 57.50 & 34.00 & 8{,}092 & 2{,}596 & 1{,}396 & 494 \\
Full INT4      & 0
  & 66.25 & 82.00 & 7{,}519 & 1{,}796 & 1{,}443 & 368
  & 56.67 & 36.00 & 8{,}469 & 2{,}612 & 1{,}508 & 501 \\
Top 10\%       & 19/22
  & 57.08 & \textbf{84.00} & 8{,}095 & 1{,}743 & 1432 & 354
  & 46.25 & 32.00 & 8{,}519 & 2{,}658 & 1491 & 509 \\
Top 20\%       & 39/44
  & 57.50 & 79.33 & 8{,}033 & 1{,}869 & 1439 & 371
  & \textbf{49.17} & 32.67 & 8{,}392 & 2{,}667 & 1412 & 506 \\
Top 30\%       & 58/67
  & \textbf{59.17} & 79.33 & 7{,}996 & 1{,}868 & 1298 & 364
  & 46.67 & 32.00 & 7{,}863 & 2{,}744 & 1440 & 507 \\
Top 40\%       & 78/89
  & 55.83 & 77.33 & 8{,}092 & 1{,}964 & 1398 & 385
  & 44.17 & 33.33 & 7{,}865 & 2{,}696 & 1501 & 499 \\
\midrule
\multicolumn{2}{l}{\textit{Best vs.\ FP16}}
  & $+$3.17\,pp & $+$12.00\,pp & $+$8.2\% & $-$9.7\% & --- & $-$9.5\%
  & $+$35.84\,pp & $-$0.67\,pp & $-$2.7\% & $+$2.4\% & --- & $+$3.0\% \\
\bottomrule
\end{tabular}%
}
\end{table*}
\begin{table*}[h]
\centering
\footnotesize
\caption{Selective compression: accuracy (\%), energy (J/query), output tokens, and power (W) --- MuSiQue (held-out evaluation split). \textbf{Bold} = best accuracy per model. Shaded row = FP16 reference.}
\label{tab:selcomp_MuSiQue}
\resizebox{\textwidth}{!}{%
\begin{tabular}{@{}llrrrr rrrr@{}}
\toprule
& & \multicolumn{4}{c}{\textbf{DeepSeek-R1-Distill-Qwen-7B}} &
    \multicolumn{4}{c}{\textbf{DeepSeek-R1-Distill-Llama-8B}} \\
\cmidrule(lr){3-6} \cmidrule(lr){7-10}
Condition & Prot.
  & \multicolumn{1}{c}{Acc (\%)}
  & \multicolumn{1}{c}{Energy (J)}
  & \multicolumn{1}{c}{Tokens}
  & \multicolumn{1}{c}{Power (W)}
  & \multicolumn{1}{c}{Acc (\%)}
  & \multicolumn{1}{c}{Energy (J)}
  & \multicolumn{1}{c}{Tokens}
  & \multicolumn{1}{c}{Power (W)} \\
\midrule
\rowcolor{gray!15}
FP16 Reference & all
  & 7.87 & 2{,}552 & 462 & 181.2
  & 14.31 & 2{,}658 & 475 & 165.4 \\
Full INT4      & 0
  & \textbf{8.00} & 2{,}465 & 448 & 147.9
  & \textbf{14.51} & 2{,}543 & 471 & 138.7 \\
Top 10\%       & 19/22
  & 6.53 & 2{,}492 & 462 & 147.7
  & 11.92 & 2{,}736 & 475 & 141.3 \\
Top 20\%       & 39/44
  & 5.27 & 2{,}430 & 454 & 148.3
  & 13.42 & 2{,}603 & 478 & 139.4 \\
Top 30\%       & 58/67
  & \textbf{8.71} & 2{,}450 & 464 & 154.6
  & 12.88 & 2{,}592 & 481 & 140.1 \\
Top 40\%       & 78/89
  & 6.40 & 2{,}463 & 461 & 155.1
  & 13.26 & 2{,}783 & 481 & 141.3 \\
\midrule
\multicolumn{2}{l}{\textit{Best vs.\ FP16}}
  & $+$0.84\,pp & $-3.9\%$ & --- & ---
  & $+$0.21\,pp & $-4.3\%$ & --- & --- \\
\bottomrule
\end{tabular}%
}
\end{table*}

\begin{enumerate}

\item FOLIO: First-Order Logic Inference

For R1-Llama-8B, Top-10\% and Top-30\% both achieve 36.67\% 
($+$3.33\,pp over FP16 at 33.33\%), matching Full INT4 at 2{,}562--2{,}633\,J. Selective compression provides no additional 
benefit over Full INT4 for Llama-8B on FOLIO, consistent with the calibration sweep showing zero vulnerability across all 224 modules, there are no vulnerable circuits to protect. For R1-Qwen-7B, selective compression degrades below Full INT4 at low $K$ (Top-10\%: 38.67\%, $-$5.33\,pp vs.\ FP16 at 44.00\%), recovering only at Top-40\% (43.33\%), which matches Full INT4. The vulnerability ranking derived on the calibration split does 
not generalize beneficially to the evaluation split for this model-task combination at low protection levels.

\item GSM8K: Grade-School Arithmetic

For R1-Llama-8B, Full INT4 degrades substantially to 70.00\% 
($-$9.33\,pp vs.\ FP16 at 79.33\%) while increasing energy to 
6{,}197\,J ($+$43.6\%) due to longer reasoning chains (1{,}092 vs.\ 771 tokens). Selective compression partially recovers 
accuracy: Top-40\% reaches 78.67\% ($-$0.67\,pp vs.\ FP16) at 6{,}223\,J, and Top-20\% achieves 78.00\% ($-$1.33\,pp) at lower energy (6{,}082\,J). Neither fully recovers FP16 accuracy while matching FP16 energy. For R1-Qwen-7B, all conditions cluster within 2\,pp of FP16 
(94.00\%), with Full INT4 (93.33\%) and Top-40\% (93.33\%) the best INT4 conditions. Selective compression provides no meaningful benefit when Full INT4 already achieves near-ceiling 
accuracy.

\item  MATH-500: Competition Mathematics

On MATH-500, selective compression provides incremental improvements 
over Full INT4 for both models. For R1-Qwen-7B, Top-30\% protection 
achieves 59.17\% ($+$3.17\,pp over FP16 at 66.25\%), with energy 
stable across protection levels (7{,}996--8{,}095\,J). Top-10\% and 
Top-20\% achieve 57.08\% and 57.50\% respectively, below both FP16 
and Full INT4 (both 66.25\%), showing that lower protection levels 
do not recover accuracy on this benchmark. For R1-Llama-8B, Top-20\% 
achieves the best result at 49.17\% ($+$35.84\,pp over FP16 at 
13.33\% and $+$5.42\,pp over Full INT4 at 43.75\%), confirming that 
selective compression substantially recovers accuracy for this model. 
The large gap between selective compression and FP16 on Llama-8B 
is consistent with the grouped quantization finding that INT4 
already outperforms FP16 on this benchmark --- the Llama-8B 
architecture benefits from the regularizing effect of quantization 
noise on competition mathematics, and selective compression refines 
which layers are compressed while preserving this benefit.

\item  ProofWriter: Formal Deductive Reasoning

For R1-Qwen-7B, selective compression achieves its strongest 
result across all benchmarks. Top-10\% (19 of 196 pairs) reaches 
84.00\%, surpassing both FP16 (72.00\%, $+$12.00\,pp) and Full 
INT4 (82.00\%, $+$2.00\,pp) while reducing energy to 1{,}743\,J 
vs.\ 1{,}931\,J for FP16 ($-$9.7\%) and generating fewer tokens 
(354 vs.\ 391). This is a genuine Pareto improvement on held-out 
data: better accuracy \emph{and} lower energy than FP16. Higher 
$K$ values degrade monotonically (Top-40\%: 77.33\%), confirming 
that protecting only the most vulnerable pairs is optimal. For R1-Llama-8B, selective compression \emph{degrades} below 
Full INT4 (36.00\%) across all $K$ levels (Top-10\%: 32.00\%, 
Top-40\%: 33.33\%), all below FP16 (34.00\%). The vulnerability 
ranking from the calibration sweep does not generalize to the 
evaluation split for Llama-8B on ProofWriter, likely because 
the low per-module vulnerability scores derive from different samples than the evaluation set.

\item MuSiQue: Multi-Hop Question Answering

MuSiQue remains intractable under all compression conditions for both models. For R1-Qwen-7B, the best selective result is Top-30\% 
at 8.71\% F1 ($+$0.84\,pp over FP16 at 7.87\%), within sampling variability. For R1-Llama-8B, Full INT4 achieves 14.51\% ($+$0.21\,pp over FP16 at 14.31\%), and selective conditions 
degrade below Full INT4 (Top-10\%: 11.92\%, $-$2.39\,pp vs.\ FP16). No selective protection strategy improves MuSiQue accuracy reliably for either model, confirming that multi-hop knowledge retrieval is uniformly sensitive to quantization at this scale and FP16 precision is necessary for this task type.

\end{enumerate}

\subsubsection{Summary}
Selective compression provides meaningful accuracy-energy gains 
when per-module vulnerability scores are high on the calibration 
split (ProofWriter Qwen-7B: Top-10\% achieves $+$12.00\,pp over 
FP16 at $-$9.7\% energy). When Full INT4 already matches FP16 
(GSM8K Qwen-7B, FOLIO Llama-8B) or vulnerability rankings do not 
generalize (ProofWriter Llama-8B, FOLIO Qwen-7B at low $K$), 
selective compression adds no benefit. Task type and per-module 
vulnerability magnitude should drive method selection. An ablation study comparing vulnerability-guided Top-$K$\% 
selection against Random-$K$\% and Bottom-$K$\% baselines 
is presented in Appendix~\ref{app:ablation}.

\subsection{Discussion}

Attention projections are more critical for competition mathematics: 
INT4+Attention achieves the best INT4 accuracy on MATH-500 for both 
models ($+$2.50\,pp over FP16 for Llama-8B; $+$3.75\,pp for 
Qwen-7B), while Full INT4 performs comparably to INT4+Attention on 
GSM8K where accuracy is already near ceiling. For logical inference, 
Full INT4 surprisingly outperforms selective group protection on 
ProofWriter for both models, with Qwen-7B Full INT4 reaching 77.08\% 
vs.\ FP16 at 69.58\% ($+$7.50\,pp), suggesting that uniform 
quantization acts as implicit regularization for formal deduction 
tasks. This reflects the mechanistic roles of these 
components~\cite{geva2021transformer}: attention implements 
token-to-token routing critical for tracking arithmetic variables; 
the ProofWriter finding suggests that MLP knowledge retrieval for 
logical rules is robust to quantization noise at the group level. 
Practitioners should select the protection strategy based on target 
task type rather than applying a universal policy.

\section{Conclusion}
We present a reasoning-aware compression framework that identifies 
vulnerable circuits in LRMs via perturbation sweep across 196--224 
(layer, projection) pairs on a held-out calibration split, and 
selectively restores them to FP16 to achieve accuracy-energy 
trade-offs validated on held-out evaluation data.

Three findings are directly actionable for deployment. First, INT4 
quantization can \emph{increase} total energy in LRMs due to 
reasoning chain extension, on R1-Llama-8B GSM8K, Full INT4 
generates 22.8\% more output tokens than FP16, converting a 25\% 
power reduction into a 22.9\% energy increase. Per-query energy 
measurement via hardware sampling is therefore essential, not 
optional. Second, vulnerability is task-dependent: attention 
projections are more critical for mathematical reasoning, while 
both attention and MLP projections contribute to logical inference 
depending on architecture. No universal protection strategy is 
optimal across all task types. Third, selective compression reaches 
Pareto-optimal operating points inaccessible to uniform methods: 
R1-Qwen-7B Top-10\% on ProofWriter achieves 84.00\% accuracy 
($+$12.00\,pp over FP16) while reducing energy by 9.7\%, validated 
on a held-out evaluation split.

Future work should explore low-rank approximation of restored layers 
to reduce memory overhead, task-adaptive vulnerability rankings that 
generalize across datasets, and evaluation on 30B--70B models where 
energy costs are most severe. 
%An anonymized codebase is available upon request to the program chairs.

\section*{Limitations}

Several limitations of this study should be acknowledged.

All experiments were conducted on a single GPU type (Tesla V100-SXM2 32\,GB) on PSC Bridges-2. Energy measurements 
and the magnitude of the chain-length extension effect may differ on other hardware. Replication on A100 or H100 hardware remains an important direction for future work.

The 30-sample evaluation size is sufficient for benchmarks with moderate accuracy (FOLIO, GSM8K, MATH-500, ProofWriter) but insufficient for reliable statistical 
inference on near-zero accuracy conditions such as MuSiQue under compression. At these accuracy levels, a single correct 
answer changes reported accuracy by 3.33\,pp, making fine-grained comparisons between conditions unreliable. The reported MuSiQue differences between the conditions should be interpreted as directional trends rather than as precise estimates.

This study evaluates only post-training compression methods. Training-aware methods such as quantization-aware training (QAT) or structured pruning with fine-tuning may 
recover more accuracy at comparable compression ratios, but require access to training data and significant additional 
compute.

\section*{Ethics Statement}
This work evaluates compression of publicly available language models on publicly available benchmarks. No human subjects were involved. The goal of reducing energy consumption supports the use of environmentally sustainable AI deployment.

\bibliography{references}

\newpage
\appendix
\onecolumn

\begin{center}
{\Large \textbf{Appendix}}
\end{center}

\vspace{4pt}

\section{Perturbation Sweep Vulnerability Heatmaps}
\label{app:heatmaps}

\begin{figure}[h]
\centering
\footnotesize
\includegraphics[width=\textwidth]{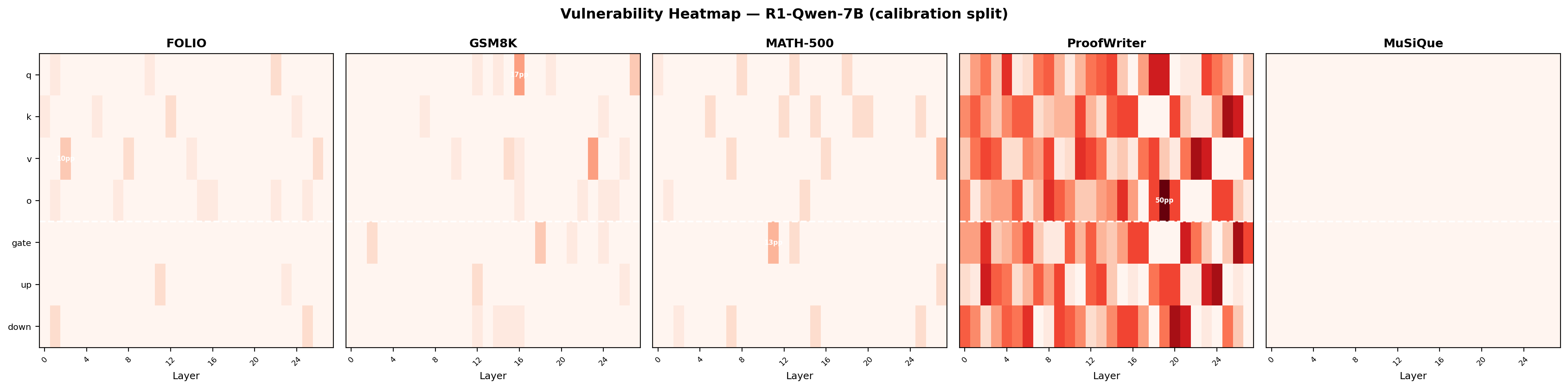}
\caption{Vulnerability heatmap for R1-Qwen-7B all benchmark. Rows = projection types; columns = layer indices (0--27).}
\label{fig:heatmap_qwen7b}
\end{figure}

\begin{figure}[h]
\centering
\includegraphics[width=\textwidth]{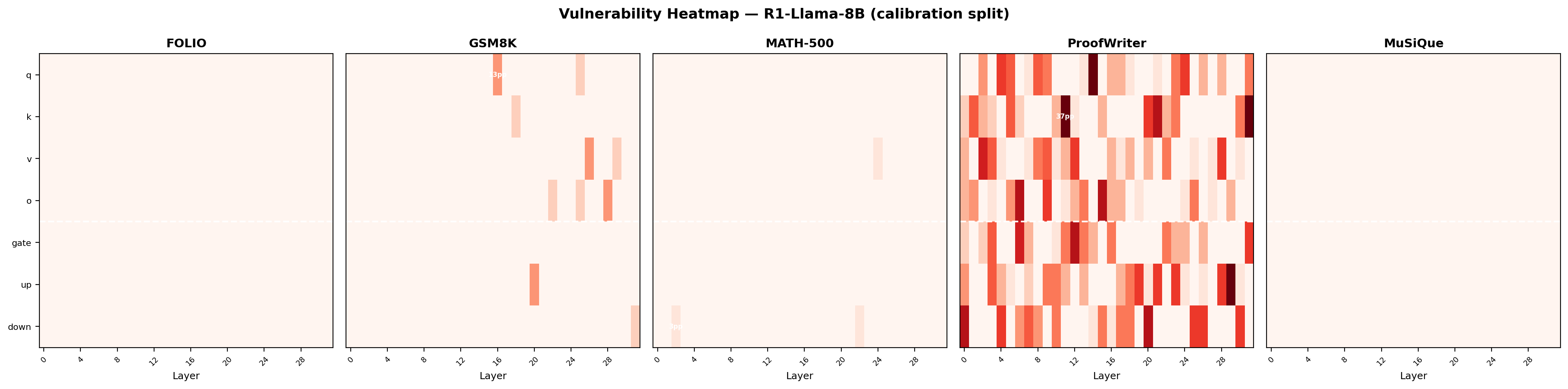}
\caption{Vulnerability heatmap for R1-Llama-8B for all benchmarks. Rows = projection types; 
columns = layer indices (0-31). The sparser pattern 
relative to R1-Qwen-7B is consistent with Full INT4 
frequently matching FP16 accuracy on Llama-8B.}
\label{fig:heatmap_llama8b}
\end{figure}

Figures~\ref{fig:heatmap_qwen7b} and~\ref{fig:heatmap_llama8b} 
show the vulnerability scores heatmaps per-module for the two models. Each cell represents the accuracy drop (pp) when that (layer, projection) pair is individually quantized to INT4. White cells indicate zero accuracy drop, safe to compress. Darker cells identify the most vulnerable circuits prioritized for FP16 restoration in selective compression.

\begin{table}[h]
\centering
\footnotesize
\caption{Per-module vulnerability summary from calibration-split
         perturbation sweep. Max drop = accuracy drop of most
         vulnerable (layer, projection) pair. Safe = pairs with
         zero or negative drop.}
\label{tab:vuln_summary}
\setlength{\tabcolsep}{4pt}
\begin{tabular}{@{}llccc@{}}
\toprule
Model & Dataset & Baseline & Max Drop & Safe Pairs \\
\midrule
R1-Llama-8B & FOLIO & 40.0\% & $+$0.0\,pp & 224/224 \\
 & GSM8K & 53.3\% & $+$13.3\,pp & 214/224 \\
 & MATH-500 & 26.7\% & $+$3.3\,pp & 221/224 \\
 & ProofWriter & 56.7\% & $+$36.7\,pp & 104/224 \\
 & MuSiQue & 0.0\% & $+$0.0\,pp & 224/224 \\
\midrule
R1-Qwen-7B & FOLIO & 43.3\% & $+$10.0\,pp & 175/196 \\
 & GSM8K & 50.0\% & $+$16.7\,pp & 170/196 \\
 & MATH-500 & 40.0\% & $+$13.3\,pp & 174/196 \\
 & ProofWriter & 76.7\% & $+$50.0\,pp & 28/196 \\
 & MuSiQue & 0.0\% & $+$0.0\,pp & 196/196 \\
\bottomrule
\end{tabular}
\end{table}

\section{Top-K vs.\ Random-K vs.\ Bottom-K Ablation}
\label{app:ablation}

Table~\ref{tab:ablation} compares vulnerability-guided (Top-$K$\%), random (Random-$K$\%, mean of 3 seeds), and least-vulnerable (Bottom-$K$\%) module protection strategies across four datasets and two models.

\begin{table}[h]
\centering
\footnotesize
\caption{Top-$K$\% vs.\ Random-$K$\% vs.\ Bottom-$K$\% accuracy (\%). Random-$K$\% is mean of 3 seeds. Full INT4 baseline shown for reference.}
\label{tab:ablation}
\setlength{\tabcolsep}{4pt}
\begin{tabular}{@{}llccc@{}}
\toprule
Dataset & $K$\% & Top-$K$ & Random-$K$ & Bottom-$K$ \\
\midrule
\multicolumn{5}{l}{\textit{R1-Llama-8B} \quad Full INT4 = 37.92\%} \\
FOLIO & 10 & 38.33 & 37.64 & 40.00 \\
FOLIO & 20 & 38.33 & 38.61 & 40.83 \\
FOLIO & 30 & 38.33 & 14.72 & 13.33 \\
FOLIO & 40 & 17.50 & 15.42 &  9.17 \\
\midrule
\multicolumn{5}{l}{\textit{R1-Llama-8B} \quad Full INT4 = 73.33\%} \\
GSM8K & 10 & 68.75 & 73.33 & 74.58 \\
GSM8K & 20 & 75.42 & 75.97 & 71.25 \\
GSM8K & 30 & 70.00 & 73.89 & 71.25 \\
GSM8K & 40 & 74.58 & 74.86 & 80.42 \\
\midrule
\multicolumn{5}{l}{\textit{R1-Llama-8B} \quad Full INT4 = 55.83\%} \\
MATH-500 & 10 & 58.33 & 55.69 & 56.25 \\
MATH-500 & 20 & 59.58 & 56.67 & 53.75 \\
MATH-500 & 30 & 58.75 & 55.42 & 56.25 \\
% MATH-500 & 40 & $\dagger$ & $\dagger$ & $\dagger$ \\
\midrule
\multicolumn{5}{l}{\textit{R1-Llama-8B} \quad Full INT4 = 35.00\%} \\
ProofWriter & 10 & 32.08 & 34.17 & 32.92 \\
ProofWriter & 20 & 36.67 & 50.83 & 54.17 \\
ProofWriter & 30 & 53.75 & 51.94 & 56.25 \\
ProofWriter & 40 & 56.67 & 54.31 & 51.25 \\
\midrule
\multicolumn{5}{l}{\textit{R1-Qwen-7B} \quad Full INT4 = 30.42\%} \\
FOLIO & 10 & 30.42 & 29.58 & 28.33 \\
FOLIO & 20 & 32.08 & 29.44 & 30.00 \\
FOLIO & 30 & 31.25 & 32.50 & 31.67 \\
FOLIO & 40 & 43.75 & 45.42 & 45.83 \\
\midrule
\multicolumn{5}{l}{\textit{R1-Qwen-7B} \quad Full INT4 = 92.50\%} \\
GSM8K & 10 & 92.92 & 93.06 & 91.67 \\
GSM8K & 20 & 92.50 & 92.50 & 92.08 \\
GSM8K & 30 & 91.67 & 92.08 & 92.08 \\
GSM8K & 40 & 93.75 & 91.67 & 92.08 \\
\midrule
\multicolumn{5}{l}{\textit{R1-Qwen-7B} \quad Full INT4 = 64.58\%} \\
MATH-500 & 10 & 66.67 & 65.69 & 62.92 \\
MATH-500 & 20 & 66.25 & 65.42 & 66.25 \\
MATH-500 & 30 & 64.58 & 67.22 & 67.08 \\
% MATH-500 & 40 & 67.92 & $\dagger$ & $\dagger$ \\
\midrule
\multicolumn{5}{l}{\textit{R1-Qwen-7B} \quad Full INT4 = 77.08\%} \\
ProofWriter & 10 & 76.67 & 75.42 & 72.08 \\
ProofWriter & 20 & 76.67 & 76.81 & 73.33 \\
ProofWriter & 30 & 76.25 & 75.14 & 72.08 \\
ProofWriter & 40 & 70.42 & 76.67 & 70.83 \\
\bottomrule
\end{tabular}
\end{table}

Results vary by dataset and model. On FOLIO and GSM8K, all three 
strategies perform within one standard deviation at $K{=}10$--$20\%$, 
and all degrade similarly at $K{=}30$--$40\%$, suggesting the 
vulnerability ranking provides limited discriminative signal when 
protecting large module fractions on these tasks. On MATH-500, 
Top-$K$\% consistently outperforms Random-$K$\% and Bottom-$K$\% 
at $K{=}10$--$20\%$ for both models, For instance, Qwen-7B Top-10\%: 
66.67\% vs.\ Random-10\%: 65.69\% vs.\ Bottom-10\%: 62.92\%); 
providing clearer evidence that the vulnerability ranking captures 
genuine module sensitivity for competition mathematics. On 
ProofWriter Llama-8B, Random-$K$\% and Bottom-$K$\% outperform 
Top-$K$\% at low $K$, suggesting the calibration-split ranking 
does not fully generalize to the evaluation split for this 
model-task combination. We interpret these mixed results as 
indicating that the vulnerability ranking is most informative 
when per-module sensitivity scores are high and consistent across 
calibration samples, a condition met on MATH-500 and ProofWriter 
Qwen-7B but not on FOLIO or GSM8K where most modules show near-zero 
vulnerability.

\section{Statistical Significance Tests}
\label{app:significance}

Table~\ref{tab:significance} reports two-sample $t$-test $p$-values 
comparing each INT4 condition against the FP16 baseline using the 
held-out evaluation split results ($n{=}5$ experiments). No accuracy 
difference reaches significance at $p{<}0.05$, reflecting the 
limited statistical power of 5 experiments at 30 samples each. 
Energy differences reach significance in 9 of 30 comparisons, 
confirming that the energy findings are robust despite small 
sample sizes.

\begin{table}[h]
\centering
\footnotesize
\caption{Two-sample $t$-test $p$-values for accuracy and energy 
         differences between each INT4 condition and FP16 baseline 
         ($n{=}5$ experiments, held-out split).
         $^{***}p{<}0.001$, $^{**}p{<}0.01$, $^*p{<}0.05$, 
         $^\dagger p{<}0.10$, n.s.\ not significant.}
\label{tab:significance}
\setlength{\tabcolsep}{4pt}
\begin{tabular}{@{}llcrcrc@{}}
\toprule
Model & Dataset & Condition & $\Delta$Acc (pp) & Sig. & $\Delta$E (J) & Sig. \\
\midrule
R1-Qwen-7B & FOLIO & Full INT4     & $-$1.67 & n.s. & $-$153 & $^\dagger$ \\
R1-Qwen-7B & FOLIO & INT4+Attn     & $-$0.83 & n.s. & $-$117 & n.s. \\
R1-Qwen-7B & FOLIO & INT4+MLP      & $-$2.92 & n.s. & $+$10  & n.s. \\
\midrule
R1-Qwen-7B & GSM8K & Full INT4     & $-$0.42 & n.s. & $-$19  & n.s. \\
R1-Qwen-7B & GSM8K & INT4+Attn     & $-$1.25 & n.s. & $-$187 & n.s. \\
R1-Qwen-7B & GSM8K & INT4+MLP      & $-$0.42 & n.s. & $+$255 & n.s. \\
\midrule
R1-Qwen-7B & MATH-500 & Full INT4  &  $0.00$ & n.s. & $-$415 & n.s. \\
R1-Qwen-7B & MATH-500 & INT4+Attn  & $+$3.75 & n.s. & $-$561 & n.s. \\
R1-Qwen-7B & MATH-500 & INT4+MLP   & $-$0.83 & n.s. & $+$209 & n.s. \\
\midrule
R1-Qwen-7B & ProofWriter & Full INT4 & $+$7.50 & n.s. & $-$131 & $^\dagger$ \\
R1-Qwen-7B & ProofWriter & INT4+Attn & $+$7.08 & n.s. & $-$214 & $^*$ \\
R1-Qwen-7B & ProofWriter & INT4+MLP  & $+$2.08 & n.s. & $+$16  & n.s. \\
\midrule
R1-Qwen-7B & MuSiQue & Full INT4   & $+$0.16 & n.s. & $-$110 & $^*$ \\
R1-Qwen-7B & MuSiQue & INT4+Attn   & $-$0.03 & n.s. & $-$94  & $^*$ \\
R1-Qwen-7B & MuSiQue & INT4+MLP    & $-$1.19 & n.s. & $+$96  & $^*$ \\
\midrule
R1-Llama-8B & FOLIO & Full INT4    & $+$3.75 & n.s. & $+$65  & $^\dagger$ \\
R1-Llama-8B & FOLIO & INT4+Attn    & $+$2.92 & n.s. & $-$110 & $^*$ \\
R1-Llama-8B & FOLIO & INT4+MLP     & $+$2.08 & n.s. & $+$240 & $^{***}$ \\
\midrule
R1-Llama-8B & GSM8K & Full INT4    & $-$6.25 & n.s. & $+$1{,}135 & $^*$ \\
R1-Llama-8B & GSM8K & INT4+Attn    & $-$1.25 & n.s. & $+$508 & n.s. \\
R1-Llama-8B & GSM8K & INT4+MLP     & $-$3.33 & n.s. & $+$1{,}611 & $^{**}$ \\
\midrule
R1-Llama-8B & MATH-500 & Full INT4 & $-$0.83 & n.s. & $+$377 & n.s. \\
R1-Llama-8B & MATH-500 & INT4+Attn & $+$2.50 & n.s. & $-$67  & n.s. \\
R1-Llama-8B & MATH-500 & INT4+MLP  & $+$1.25 & n.s. & $+$610 & n.s. \\
\midrule
R1-Llama-8B & ProofWriter & Full INT4 & $+$2.50 & n.s. & $+$52  & n.s. \\
R1-Llama-8B & ProofWriter & INT4+Attn & $+$0.83 & n.s. & $-$89  & $^*$ \\
R1-Llama-8B & ProofWriter & INT4+MLP  & $-$0.83 & n.s. & $+$367 & $^{***}$ \\
\midrule
R1-Llama-8B & MuSiQue & Full INT4  & $-$0.28 & n.s. & $-$72  & $^\dagger$ \\
R1-Llama-8B & MuSiQue & INT4+Attn  & $-$1.95 & n.s. & $-$119 & $^*$ \\
R1-Llama-8B & MuSiQue & INT4+MLP   & $-$0.85 & n.s. & $+$52  & n.s. \\
\bottomrule
\end{tabular}
\end{table}

\section{Use of LLMs}

Claude (Anthropic) was used to assist with the the polishing of small fractions of sentences during this project. All experimental results, scientific claims, and the final manuscript content were verified and approved by the authors.

\end{document}